%% file: concausality-frame.tex
\pdfoutput=1
\documentclass[11pt]{article}
\usepackage[preprint]{acl}

\usepackage{concausality-frame}

\graphicspath{{./concausality-figures}}

\newcommand{\dataset}{Countercausal News Corpus}

\begin{document}
\input{concausality-pre}
\input{concausality-part1}
\input{concausality-part2}
\input{concausality-part3}
\input{concausality-part4}
\input{concausality-part5}
\input{concausality-sum}

\input{concausality-limitations}
\input{concausality-ethics}

\bibliography{concausality-lit}

\clearpage
\appendix
\input{concausality-appendix}
\end{document}

%% file: concausality-pre.tex
\title{Investigating Counterclaims in Causality Extraction from Text}

\author{
Tim Hagen \\
University of Kassel\\and hessian.AI\And
Niklas Deckers \\
University of Kassel\\and hessian.AI\\\And
Felix Wolter \\
Leipzig University \\\AND
Harrisen Scells \\
University of Tübingen\And
Martin Potthast \\
University of Kassel,\\hessian.AI, and ScaDS.AI
}

\maketitle

\begin{abstract}
Many causal claims, such as ``sugar causes hyperactivity,'' are disputed or outdated. Yet research on causality extraction from text has almost entirely neglected counterclaims of causation. To close this gap, we conduct a thorough literature review of causality extraction, compile an extensive inventory of linguistic realizations of countercausal claims, and develop rigorous annotation guidelines that explicitly incorporate countercausal language. We also highlight how counterclaims of causation are an integral part of causal reasoning. Based on our guidelines, we construct a new dataset comprising 1028~causal claims, 952~counterclaims, and 1435~uncausal statements, achieving substantial inter-annotator agreement (Cohen's~$\kappa\,{=}\,$0.74). In our experiments, state-of-the-art models trained solely on causal claims misclassify counterclaims more than 10~times as often as models trained on our dataset.
\end{abstract}

%% file: concausality-part1.tex
\section{Introduction}

Knowledge about causality is predominantly communicated through language. The scientific literature, textbooks, news articles, and the web abound with statements that assert causation. Extracting such causal claims from text and building causal knowledge graphs \cite{hassanzadeh:2024, arsenyan:2023} supports decision-making in domains including healthcare \cite{zhao:2022a, mihaila2013}, finance \cite{sakaji2023}, and risk management \cite{ravivanpong2022}. And since large language models struggle with reliable causal reasoning \cite{joshi:2024}, causal graphs are important for causal question-answering systems as well \cite{bondarenko2022, hassanzadeh:2019}.

\bsfigure{countercausality-illustration2}{Countercausal claims such as ``A~does not cause~B'' are currently neglected in causality extraction from text. Our dataset is the first that includes them.}

\input{table-causality-extraction-pipeline-examples}

However, many causal claims are subject to controversial discourse, including claims about climate\colbreak{} change, vaccination, or parenting. Claims once assumed to be true, such as sugar consumption causing hyperactivity in children, may later be disproven \citep{wolraich:1995}, yet public belief in them often persists for a long time. It is therefore surprising that research on causality extraction from text has almost entirely neglected counterclaims (see Figure~\ref{countercausality-illustration2} and Section~\ref{related-work}). Our literature review shows that this omission contrasts with causal reasoning on incomplete knowledge, in which countercausal claims are essential.

To address this limitation, we make three contributions:
\Ni
We derive annotation guidelines for a wide range of causal and countercausal claims and explain why both are important for causal reasoning (Section~\ref{sec:countercausality}).
\Nii
We compile a dataset of 3415~causal claims, countercausal claims, and uncausal statements, enabling the training and evaluation of better causality extraction models (Section~\ref{sec:corpus}).
\Niii
We evaluate the impact of the absence and presence of countercausal claims during training on the effectiveness of state-of-the-art causality extractors.%
\footnote{Code, data: \href{https://github.com/webis-de/arxiv-countercausality}{github.com/webis-de/arxiv-countercausality}}

%% file: table-causality-extraction-pipeline-examples.tex
\begin{table*}
\centering
\small
\tabcolsep=4.927pt%
\renewcommand{\arraystretch}{0.85}
\belowrulesep=\dimexpr\arraystretch \belowrulesep\relax%
\aboverulesep=\dimexpr\arraystretch \aboverulesep\relax%
\begin{tabular}{@{}llll@{}}
\toprule
\bfseries Pipeline step & \multicolumn{3}{@{}c@{}}{\bfseries Example} \\
\cmidrule(l@{\tabcolsep}){2-4}
&
Not permitting bars caused a protest &
No person was left stranded by the strike &
We are not on strike \\
\midrule

\bfseries Causality detection: &
causal, explicit &
countercausal, implicit &
uncausal \\\noalign{\vspace{.1em}}

\bfseries Event extraction: &
$A$: ``not permitting bars'' &
$C$: ``person was left stranded'' \\

&
$B$: ``a protest'' &
$D$: ``the strike'' \\\noalign{\vspace{.1em}}

\bfseries Causality identific.: &
$A \to B$ &
$D \nrightarrow C$ \\
\bottomrule
\end{tabular}\vspace{-1ex}
\caption{Causality extraction is often divided into a pipeline of causality detection, event extraction, and causality identification. So far, countercausal claims have not been modelled, but have been classified as noncausal.}
\label{table-causality-extraction-pipeline-examples}
\end{table*}

%% file: concausality-part2.tex
\section{Related Work}
\label{related-work}

\paragraph{Causality Extraction from Text}
The extraction of cause--effect relationships from text differs from many other knowledge extraction tasks, since causality is often expressed implicitly without signal words like `because' \citep{prasad:2008,yang:2022b} and with a high linguistic diversity \cite{hidey:2016}. Besides end-to-end approaches \cite{gao2022a,dasgupta:2022,zheng:2017}, the task is therefore commonly addressed using an extraction pipeline (\citet{tan2023b}; Table~\ref{table-causality-extraction-pipeline-examples}):
\Ni
causality detection, where given a text, it is determined if it contains causal information;
\Nii
event extraction, where given a text containing causal information, text spans are identified as candidates for causally related events (e.g., using sequence labeling \cite{li2021}); and
\Niii
causality identification, where given a text and candidate events it is determined if the text asserts them to be causally related (e.g., by classifying ordered pairs of candidates \cite{liu:2020}).

\input{table-causality-corpora}

\paragraph{Datasets}
Table~\ref{table-causality-corpora} lists the available causality extraction datasets. They typically consist of English sentences. The Causal News Corpus~(CNC) and the CNCv2 are sampled from English newspaper articles from India \cite{tan2022a, tan2023a}, which makes annotation challenging if one lacks cultural context. Most corpora are sampled from US~news; the FinCausal corpora use financial reports \cite{mariko:2020,moreno-sandoval:2023}; Task~8 at the Sem\-Eval corpora switched from web texts via news to medical causal claims discussed on social media \cite{khetan:2023}. The sentences of all prior corpora are labeled as causal~(c) or noncausal~(nc), subsuming countercausal~(cc). Our corpus is the first to divide noncausal into uncausal~(uc) and countercausal.%
\footnote{Appendix~\ref{sec:naming} is a glossary of causality terminology.}

Causal sentences are labeled with cause and effect spans in all corpora. The CNCv2 also labels signal spans (where explicit). The BECauSE corpora \cite{dunietz:2015, dunietz:2017} further label causal sentences as consequence, motivation, purpose, or inference. To increase annotator agreement, BECauSE~2.0 changes inferences to uncausal, since they may not indicate ontic causality except through abductive reasoning \cite{grivaz2010}. While there is no universally accepted labeling standard for causality \cite{xu2020}, unified labeling schemes were proposed such as CaTeRs \cite{mostafazadeh:2016}, CREST \cite{hosseini:2021}, PolitiCause \cite{corral:2024} and UniCausal \cite{tan2023b}.

\paragraph{Models for Causality Extraction}
\Citet{yang:2022b} distinguish rule-based from classic and deep learning approaches. The former use manually compiled linguistic patterns to identify explicit\colbreak{} signal words like `because' \cite{girju:2003,bui:2010,cao:2014,mirza:2014}. The latter include knowledge-oriented convolutional neural networks (K-CNNs, \citet{li2019}), which employ world knowledge to learn semantic and syntactic features, and SCITE \cite{li2021}, which applies self-attention to an LSTM to perform sequence tagging for joint extraction of cause--effect pairs. More recently, trans\-for\-mer-based models have been used \cite{khetan2021, hosseini:2021, yang:2022c}. But language models still struggle with many forms of causality, such as implicit causality \cite{takayanagi:2024} or counterfactual reasoning \cite{gendron:2024b}.

\paragraph{Countercausality}
Negations of causation hardly have been studied in causality extraction. \citet{sanchez-graillet2007} study negations of protein--protein interactions in scientific literature. \citet{mirza:2014a}, \citet{bui:2010}, and \citet{pawar:2021} consider counterclaims only through negation of causal claims, while ignoring other forms of countercausality (e.g., ``A~despite~B'' is synonymous to ``B~did not prevent~A'', but only the latter is annotated). For PolitiCause, \citet{corral:2024} ask annotators to label statements like ``The policy did not improve the situation'' as causal, conflating its semantics with ``The policy caused the non-improvement.'' \citet{cui:2024a} recently first explored defeasibility through causal counterclaims in causal reasoning on language.

\paragraph{Applications}
Causality extraction enables building causal graphs \cite{heindorf:2020,priniski:2023,zhang:2022a}, which are used for causal inference \cite{jin:2023}. Both are instrumental to automatic decision-making, e.g., in medicine \cite{nordon:2019} and finance \cite{nayak:2022}. Computational argumentation can be modeled as epistemic causality \cite{alkhatib:2023}, where premises cause an arguer to believe their conclusion holds. Causal argumentation is especially relevant to topics such as climate change \cite{allein:2025}. Causality-related queries account for~5\% of web searches \cite{bondarenko2022}. Although large language models~(LLMs) are increasingly used for question answering, they perform poorly at causal reasoning \cite{romanou:2023,joshi:2024,jin:2024}, erring and hallucinating on such questions \cite{gao:2023}.

%% file: table-causality-corpora.tex
\begin{table}
\centering
\small
\renewcommand{\tabcolsep}{3.65pt}
\renewcommand{\arraystretch}{0.88}
\belowrulesep=\dimexpr\arraystretch \belowrulesep\relax%
\aboverulesep=\dimexpr\arraystretch \aboverulesep\relax%
\newcommand{\x}{\makebox[0cm][c]{\checkmark}}
\begin{tabular}{@{}l@{}rrcccll@{}}
\toprule
\bfseries Corpus & \multicolumn{3}{@{}c@{}}{\bfseries Sentences} & \multicolumn{2}{@{}c@{}}{\bfseries\kern-0.4em Signal} &\bfseries Dom. & \bfseries Year \\% \bfseries IAA &
\cmidrule(l@{\tabcolsep}r@{\tabcolsep}){2-4}\cmidrule(l@{\tabcolsep}r@{\tabcolsep}){5-6}
\addlinespace[-0.25ex]
& \multicolumn{1}{@{}c@{}}{c} & \multicolumn{1}{@{}c@{}}{nc/uc} &              cc & e  & i &         &                                 \\%$\kappa$ &
\midrule
\multicolumn{8}{@{}l@{}}{\em Manual labeling} \\
  BECauSE                     &            400 &            800 & \color{gray} -- & \x &    & news   & \citeyear{dunietz:2015}         \\%    --   &
  BECauSE~2.0                 &         1\,803 &         3\,577 & \color{gray} -- & \x &    & news   & \citeyear{dunietz:2017}         \\%   0.91  &
  BioCause                    &            851 &             -- & \color{gray} -- & \x &    & med.   & \citeyear{mihaila2013}          \\%    --   &
  CaTeRs                      &            488 &         2\,715 & \color{gray} -- &    &    & fict.  & \citeyear{mostafazadeh:2016}    \\%    --   &
  Causal TimeBank             &            318 &         1\,418 & \color{gray} -- & \x &    & news   & \citeyear{mirza:2014a}.         \\%    --   &
  CNC                         &         1\,957 &         1\,602 & \color{gray} -- & \x & \x & news   & \citeyear{tan2022a}             \\%    --   &
  CNCv2                       &         1\,809 &         1\,606 & \color{gray} -- & \x & \x & news   & \citeyear{tan2023a}             \\%    --   &
  EventStory Line             &         1\,156 &         1\,076 & \color{gray} -- & \x & \x & news   & \citeyear{caselli:2017}         \\%    --   &
  FinCausal-20                &         2\,136 &        27\,308 & \color{gray} -- & \x & \x & fin.   & \citeyear{mariko:2020}          \\%    --   &
  FinCausal-23                &         3\,432 &             -- & \color{gray} -- & \x &    & fin.   & \citeyear{moreno-sandoval:2023} \\%    --   &
  PDTB-2                      &         8\,042 &        28\,550 & \color{gray} -- & \x & \x & news   & \citeyear{prasad:2008}          \\%    --   &
  PolitiCause                 &         5\,070 &        12\,710 & \color{gray} -- & \x &    & polit. & \citeyear{corral:2024}          \\%    --   &
  SemEval-07 T.\ 4            &            220 &            114 & \color{gray} -- & \x &    & web    & \citeyear{girju:2007}           \\%    --   &
  SemEval-10 T.\ 8            &         1\,331 &         9\,386 & \color{gray} -- & \x &    & web    & \citeyear{hendrickx:2010}       \\%    --   &
\bfseries
  CCNC (ours)                 &         1\,028 &         1\,435 &             952 & \x & \x & news   & 2025                            \\%   0.74  &
\midrule
\multicolumn{8}{@{}l@{}}{\em Automatic or assisted labeling} \\
  AltLex                      &         9\,190 &        72\,135 & \color{gray} -- & \x &    & wiki   & \citeyear{hidey:2016}           \\%    --   & % Weak supervision
 %corr2cause                  &                &                & \color{gray} -- & \x &    & synth. & \citeyear{jin:2024}             \\%    --   &
  EventCausality              &            583 &             -- & \color{gray} -- & \x &    & news   & \citeyear{do:2011}              \\%    --   &
\midrule
\multicolumn{8}{@{}l@{}}{\em Crowdsourced} \\
  SemEval-20 T.\ 5            &         2\,192 &        17\,808 & \color{gray} -- & \x &    & news   & \citeyear{yang:2020}            \\%    --   &
  SemEval-23 T.\ 8            &            597 &         5\,098 & \color{gray} -- & \x & \x & med.   & \citeyear{khetan:2023}          \\%    --   &
\midrule
\multicolumn{8}{@{}l@{}}{\em Combined corpora} \\
  UniCausal                   &        14\,903 &        43\,817 & \color{gray} -- & \x & \x & mult.  & \citeyear{tan2023b}             \\%    --   &
\bottomrule
\end{tabular}\vspace{-1ex}
\caption{Overview of causality extraction corpora. Sentences are either causal~(c), noncausal~(nc) or uncausal~(uc), or countercausal~(cc). Corpora contain explicit~(e) or implicit~(i) causality signals.}
\label{table-causality-corpora}
\end{table}

%% file: concausality-part3.tex
\section{Countercausality in Language and Reasoning}
\label{sec:countercausality}

In this section, we derive annotation guidelines that integrate countercausal language from prior work on causal language. We then discuss how countercausal knowledge is inherent to causal reasoning.

\subsection{Annotating Countercausal Claims}

Annotating causality in natural language has been shown to be challenging since it is not a syntactic but a semantic property of text. Although annotators often have an intuitive sense of causality, lack of clear guidance typically leads to low inter-annotator agreement \cite{grivaz2010}. There is no universally accepted definition of causality in text. Therefore, even carefully designed annotation guidelines do not fully eliminate ambiguity and uncertainty \cite{dunietz:2015}. We build on \citeauthor{grivaz2010}'s causality annotation guidelines, which have also been used to annotate several datasets \cite{mihaila2013, dunietz:2017, tan2022a}, including the Causal News Corpus~v2 on which we build. These guidelines decompose causality into a small number of features and guiding questions.

\paragraph{Causality}
\Citeauthor{grivaz2010} outlines three necessary conditions for two events to be in a causal relation (Table~\ref{table-countercausality-properties}):
\Ni
temporal order: the effect cannot occur before the cause;
\Nii
counterfactuality: the effect is less likely without the cause; and
\Niii
ontological asymmetry: the effect only rarely causes the cause.%
\footnote{Cause and effect may be difficult to distinguish (chicken-or-egg dilemma) or influence each other (reciprocal causation).}
If any of these conditions is not met, the relation is at least noncausal or even uncausal. For example, ``glass is fragile because of its molecular structure'' is uncausal despite the signal word `because', since it violates ontological asymmetry \cite{rosen:2010, skow2014}. However, even taken together, these conditions are not sufficient for a statement to express causation, as they do not distinguish causation from correlation. Consider the statement ``It rained~(A) and people used umbrellas~(B), followed by a flood~(C),'' which mentions the events~A, B, and~C. The relation $B \to C$ satisfies all three necessary conditions for causality: B~occurs before~C; B~and~C are correlated through the common cause~A such that, if~A would not have been observed, the likelihood of~C increases upon observing~B; and~C does not cause~B. Yet, the relation between~B and~C is clearly not causal but arises from the confusing ellipsis.

\Citeauthor{grivaz2010} guides annotators by instructing them to perform two test exercises during annotation:
\Ni
Causal chain test: Try to construct a plausible chain of events that leads from a statement's presumed cause to its effect. For example, ``the vase broke because it was knocked over'' is causal, since knocking the vase over may cause it to fall, which in turn may cause it to impact a surface and thus break.
\Nii
Paraphrase test (called \textit{linguistic test} by \citeauthor{grivaz2010}): Can the statement be paraphrased as ``A~causes~B''?
Both tests still require prior domain knowledge and an intuition of causality, but were found to help annotators to systematically reflect on a statement.

\input{table-countercausality-properties}

\paragraph{Countercausality}
Table~\ref{table-countercausal-expressions} shows how \citeauthor{grivaz2010}'s conditions can be violated, which has been the first step in deriving our extended annotation guidelines.

Since temporal order and counterfactuality are necessary conditions for causality, a violation of either is sufficient to refute a causal relation. Violations of temporal order~(a) occur when the potential cause is stated to happen after the effect. Violations of counterfactuality deny that the potential cause increases the likelihood of the effect, for example by stating that
(b)~%
the two events co-occur independently (``A~and~B coincided''), or
(c)~%
the potential cause occurred without the effect following.
Violations of counterfactuality also include statements that express surprise about observations that contradict expectations of causality (d,\,e; e.g., ``B~happened though A~occurred'').

An exception is a violation of ontological asymmetry~(f), which indicates uncausality rather than countercausality. Ontological asymmetry must hold in neither direction: ``A~causes not~B'' is in fact a causal statement in which ``not~B'' is the effect caused by~A. A~causing~B may further be refuted by denying that~B is caused at all~(g). But since research on causality commonly assumes the principle of universal causation \cite{mccain:1997}, namely that everything that happens has a cause, this, too is uncausal. However, assuming human intuition or expertise about causality, this assumption supports \citeauthor{grivaz2010}'s causal chain test.

Analogous to the paraphrase test for causality, countercausality can be tested by attempting to rephrase a statement as ``A~does not cause~B''~(h,\,i). But negation in natural language is not limited to such a narrow scope. Countercausality may also be expressed through negations like ``It is not the case that A~causes~B'' or through purely semantic negation (``It is falsely believed that A~causes~B'').

\input{table-countercausal-expressions}

\medskip
Finally, annotating countercausality requires conceivability of the underlying causal relation between two events in the given context. This requirement mirrors the causal chain test: just as ``A~causes~B'' requires that a causal chain between~A and~B is conceivable given the text, ``A~does not cause~B'' requires that such a chain would have been conceivable without the statement. Conceivability is often implicit but may be made explicit, for example through the expressions of surprise mentioned above (e.g., ``Surprisingly, the alarm went off before he entered the building'').

Note that the categories above are a practical categorization of countercausal claims rather than an exhaustive taxonomy. Some statements may fall into mutliple categories, for instance, lack of effect can be viewed as a specific realization of denying counterfactuality. During the annotation of our dataset, all statements of countercausality fit into these categories.

\subsection{Countercausality in Causal Reasoning}
\label{sec:reasoning}

While in natural language a countercausal claim can often be formed by simply negating a causal claim, the situation is more subtle in logical reasoning. This section explains how the extraction of countercausal claims aids causal reasoning. We first review how causal reasoning is commonly performed on causal graphs and why this approach fails to capture key aspects of human reasoning. We then revisit formal logical approaches to causal reasoning and explain how they benefit from incorporating countercausal claims.

Current approaches to causal reasoning over causal graphs primarily focus on identifying causal chains \cite{blubaum:2024}. Formally, this reasoning can be interpreted as operating over a set of Horn clauses, where each event is represented as a proposition and a causal statement ``A~causes~B'' is modeled as a logical implication $A\,{\to}\,B$. A potential effect~$B$ can then be deduced from a potential cause~$A$, denoted $A\,{\mathrel{\vdash}}\,B$, if and only if a causal chain exists from~$A$ to~$B$. This form of reasoning assumes a closed-world: ``A~does not cause~B'' is taken to hold if and only if no causal chain from~$A$ to~$B$ exists, that is, $A\,{\mathrel{\nvdash}}\,B$.

The problem with this assumption is that causal knowledge extracted from text is inherently incomplete and may be contradictory. A corpus of text may both assert and refute that~$A$ causes~$B$. Such contradictions can arise from evolving knowledge, controversial topics, or from the underspecification of causes and effects in human communication. Speakers typically provide only as much information as required \citep{grice:1975}, and it may even be impossible to specify events with sufficient precision to guarantee causation \cite{russell:1912}. As a result, causal reasoning is nonmonotonic: conclusions drawn from general causal assumptions may be withdrawn when additional information becomes available (Table~\ref{table-reasoning-nonmonotonicity}). This nonmonotonicity is a core property of how humans communicate and reason about causality \citep{cummins:1995}.

Propositional logic is monotonic and therefore ill-suited for reasoning over incomplete knowledge. In particular, negating causality is not equivalent to negating an implication. Modeling ``A~does not cause~B'' as $\lnot(A\,{\to}\,B)$ would require that~$A$ occurs while~$B$ does not, although in reality both events may still cooccur without being in a causal relation. To support reasoning over incomplete knowledge, \citet{reiter:1980} introduce default logic. Default logic allows for the derivation of defeasible beliefs in the absence of contradicting evidence and has been applied to causal reasoning in prior work \citep{pearl:1988, geffner:1994, bochman:2023}. \citet{poole:1991} further introduce a prioritization of more specific defaults, which aligns well with human reasoning. However, default logic cannot explicitly represent countercausal claims. Statements such as ``if~C, then~A does not cause~B'' must instead be encoded analogously to negative causal claims. While this distinction can be abstracted away during reasoning, it is crucial for causality extraction, since negative causation is only one possible linguistic realization of a countercausal claim.

\input{table-reasoning-nonmonotonicity}

Current approaches to causal reasoning over graphs extracted from natural language typically assume that their contents are complete and consistent. These assumptions do not hold in practice. Instead, causal reasoning over extracted knowledge necessarily requires nonmonotonicity, and countercausal claims become essential because they allow more specific information to override more general causal beliefs. Extracting countercausal claims therefore provides a necessary foundation for future work on causal reasoning over causal knowledge extracted from text.

%% file: table-countercausality-properties.tex
\begin{table}%
\centering%
\small%
\setlength{\tabcolsep}{0.2em}%
\renewcommand{\arraystretch}{0.9}%
\newcommand{\group}[2]{\multirow{#1}{*}{\rotatebox[origin=c]{90}{\makebox[0pt][c]{#2}}}}%
\newcommand{\narrow}[1]{\scalebox{.85}[1.0]{#1}}%
\begin{tabular}{@{}l@{\hspace{4pt}}lp{4.945cm}@{}}
\toprule
  \multicolumn{2}{@{}l}{\textbf{Feature}} & \textbf{Examples to be checked for causality }        \\
\midrule
\group{3}{\narrow{Required}} 
   & Temporal order                       & \itshape The vase broke before the fall.              \\
   & Counterfactuality                    & \itshape He came home and his mailbox is empty.       \\
   & Ontol. asymm.                        & \itshape It is a triangle because it has three sides. \\
\midrule
\group{2}{Help}
   & Causal chain test                    & \itshape The vase broke because it fell.              \\
   & Paraphrase test                      & \itshape It is Johns birthday, and he is happy.       \\
\bottomrule
\end{tabular}\vspace{-1ex}
\caption{The necessary conditions and reflective questions to test causal language designed by \citet{grivaz2010}.}%
\label{table-countercausality-properties}%
\end{table}

%% file: table-countercausal-expressions.tex
\begin{table}
\small%
\renewcommand{\arraystretch}{.8}%
\belowrulesep=\dimexpr\arraystretch \belowrulesep\relax%
\aboverulesep=\dimexpr\arraystretch \aboverulesep\relax%
\belowbottomsep=\belowrulesep%
\newcommand{\tmp}[1]{\noalign{\vspace{3pt}\textbf{#1}\vspace{2pt}}}
\begin{tabular}{@{}>{\itshape\quad}p{\linewidth}@{}}
\toprule
\kern-1em\normalfont\bfseries Expressions of Countercausal Claims \\
\midrule

\tmp{(a) Violation of temporal order}
A happened only after B                    \\

\midrule

\tmp{(b) Violation of counterfactuality}
B is equally likely without A              \\
A and B happened coincidentally            \\
A happened; B happened independently of A  \\

\tmp{(c) Lack of effect}
A happened and B did not happen            \\
A was done in vain to achieve B            \\
A is insufficient to achieve B             \\
He is happy. He did not win the lottery.   \\

\tmp{(d) Inverse expected cause {\hfill\normalfont\footnotesize ($A\!\to\!B$ is typically expected)}}
B happened though A did not happen         \\

\tmp{(e) Usual inverse effect {\hfill\normalfont\footnotesize ($A\!\to\!\lnot B$ is typically expected)}}
B happened despite A                       \\
A did not prevent B                        \\

\midrule

\tmp{(f) Negative causation\hfill{\normalfont\footnotesize (Violates ontol. asymmetry)}}
A causes not B                             \\
A prevents B                               \\

\tmp{(g) Contradiction\hfill{\normalfont\footnotesize (Violates principle of universal causation)}}
B has no cause                             \\
B never happens                            \\

\midrule

\tmp{(h) Direct negation\hfill{\normalfont\footnotesize (Analogue to paraphrase test)}}
A does not cause B                         \\
B is caused by something else than A       \\
B is only caused by \dots (not listing A)  \\

\tmp{(i) Negated context}
It is falsely believed that A causes B     \\
It is not that A causes B                  \\

\bottomrule
\end{tabular}\vspace{-1ex}
\caption{Different ways of counterclaiming a causal relationship from $A$ to $B$ and some examples (italic).}
\label{table-countercausal-expressions}
\end{table}

%% file: table-reasoning-nonmonotonicity.tex
\begin{table}\small
\setlength{\tabcolsep}{0.236em}%
\renewcommand{\arraystretch}{.88}%
\belowrulesep=\dimexpr\arraystretch \belowrulesep\relax%
\aboverulesep=\dimexpr\arraystretch \aboverulesep\relax%
\belowbottomsep=\belowrulesep%
\begin{tabular}{@{}lll@{}}
\toprule
\textbf{Step} & \textbf{Learned information} & \textbf{Conclusion} \\
\midrule
1 & Rock thrown at window      & Window shatters \\
2 & Window is bulletproof      & \textit{[Conclusion withdrawn]} \\
3 & Window was cracked already & Window shatters \\
\bottomrule
\end{tabular}
\caption{Human reasoning is nonmonotonic: learning new facts can invalidate prior conclusions. This example represents conclusions with increasing knowledge.}
\label{table-reasoning-nonmonotonicity}
\end{table}

%% file: concausality-part4.tex
\section{The Countercausal News Corpus}
\label{sec:corpus}

This section describes the construction of the \dataset{} and the tasks it supports (similarly to \citet{tan2023a, tan2023b}).

\input{table-corpus-statistics}

\subsection{Corpus Construction}

As a starting point for our \dataset{}, we selected the Causal News Corpus~v2 (CNCv2) \cite{tan2023a}. Our rationale for reusing an existing corpus rather than building a new one is to maximize synergy and comparability with pre-existing research. We selected this corpus for its complex sentence structures and annotations that indicate instance difficulty, including whether causality is signaled explicitly or implicitly.

To augment the corpus with countercausal samples, we randomly sampled approximately half of the causal sentences in~CNCv2 and rewrote them manually. We dismissed automatic rewriting, since language models remain prone to errors when handeling negation in natural language \cite{kassner2020, ravichander2022, hossain:2022, weller2024b}. Preliminary experiments confirmed these limitations in our setting: Models paraphrased too much, failed to identify all causal statements, or tended to produce negative causal but not countercausal statements (see Appendix~\ref{sec:prompt-reformulation-test}, Table~\ref{tbl:gptexamples} for an example). Moreover, since LLM-based causality extractors will be evaluated using our dataset, generate samples using an LLM risks biasing them toward ease of LLM-based extraction. Manual rewriting enabled us to deliberately include difficult samples, such as ``It is wrongly claimed that $A$ causes $B$.''

After rewriting, two annotators annotated the entire dataset based on our annotation guidelines (Appendix~\ref{sec:annoation-guidelines}). This let us identify existing sentences that express countercausality (54 instances), and ensure that all rewritten sentences were indeed countercausal. Countercausality (``$A$~does not cause~$B$'') can easily be confused with negative causation (``$A$~causes not~$B$''), a distinction that the annotation phase helped to identify.

An pilot annotation of two annotators on a sample of 100~sentences already yielded a substantial inter-annotator agreement of Cohen's~$\kappa\,{=}\,0.69$. Through discussion, the annotators identified and resolved the following sources of disagreement and revised the annotation guidelines accordingly:
\Ni
If~$A$ and~$B$ have a common cause~$C$ and are not in a causal chain, then the text is uncausal unless~$C$ is mentioned, and,
\Nii
the \emph{purpose} relation \cite{webber:2019} can be causal if it describes the trigger (e.g., ``A is done to protest against B''; B~happens\,\textrightarrow{}\,it is not liked\,\textrightarrow{}\,A~happens) or an immediate result (e.g., ``He puts money on the bank to keep it safe''), but not if it only expresses an abstract goal (e.g., ``He moved to America to make his dreams come true'').
Similarly to \citet{dunietz:2017}, we then measured the inter-annotator agreement of our now educated annotators on a fresh sample of 300~sentences, achieving an improvement to~$\kappa\,{=}\,0.74$. Based on these results, the reannotation was then extended to the entire dataset, which allowed us to identify previously mislabeled samples (see Appendix~\ref{sec:training-data-excerpt}, Table~\ref{tbl:misannotated}).

Corpus statistics and class distributions are reported in Table~\ref{table-corpus-statistics}. We retained the original splits from CNCv2 so that models can be trained on CNCv2 and evaluated on our dataset version, and vice versa, without data leakage. For provenance, each sentence of our dataset still has its original CNCv2~ID, enabling a synopsis of causal and countercausal sentence variants. Text span annotations for causes and effects were directly transferred from~CNCv2, as the underlying spans remained largely aligned after reformulation. The flesch reading ease score of 43 indicates that the texts are difficult to read. This may make the dataset also more challenging for pre-trained language models.
The dataset is compatible with Hugging Face Datasets \cite{lhoest:2021}.

\subsection{Causality Extraction Tasks}
\label{sec:tasks}

Our dataset enables the evaluation of the three tasks of \citeauthor{tan2023b}'s causality extraction pipeline, extended to account for countercausality:

\paragraph{Task~1: Causality Detection}
Causality detection is a text classification task: Given a natural language text, does it contain causal information? This task can be extended to countercausality by modeling it as
\Ni
a ternary classification problem (causal, countercausal, or uncausal), or
\Nii
a binary classification problem with causal and countercausal as the positive class and uncausal as the negative class.
The latter has the advantage that, in practice, a causal sentence may contain a mixture of causal and countercausal information, and the causal or the countercausal label cannot be determined at the level of the whole sentence but must instead be resolved during causality identification for specific event pairs.

\paragraph{Task~2: Event Extraction}
In event extraction, the model is given a sentence (and the causal class to which it belongs), and the task is to output text spans that are plausible candidates to be in a cause--effect relation. This can be modeled as a sequence tagging problem using BIO~tags \cite{ramshaw:1995}, where, for each input token, the model predicts whether it marks the \textbf{b}eginning of a \textbf{c}ause or \textbf{e}ffect span (B-C or B-E), lies \textbf{i}nside a \textbf{c}ause or \textbf{e}ffect span (I-C or I-E), or is \textbf{o}utside both~(O) \cite{li2021}. Since this task only extracts candidate events that are subsequently classified with respect to their specific relation, it does not need to be modified to support countercausality.

\paragraph{Task~3: Causality Identification}
Given a causal sentence and a pair of candidate events, the task is to determine whether the events in a causal relation. \Citet{tan2023b}, for example, add four special tokens to the tokenizer ({\tokenfont \textlangle ARG0\textrangle}, {\tokenfont \textlangle /ARG0\textrangle}, {\tokenfont \textlangle ARG1\textrangle}, and {\tokenfont \textlangle /ARG1\textrangle}) to mark candidate events and learn a binary classifier that determines whether the event marked by {\tokenfont ARG0} causes the event marked by {\tokenfont ARG1}. To extend this task to countercausality, the formulation can be generalized from binary classification (causal vs.\ noncausal) to ternary classification (causal, countercausal, or uncausal).

%% file: table-corpus-statistics.tex
\begin{table*}
\centering
\small
\tabcolsep=1.5pt%
\begin{tabular}[t]{@{}lrrr@{}}
\addlinespace[-2ex]
(a)\\
\toprule
\bfseries Class & \multicolumn{2}{@{}c@{}}{\bfseries Splits} & \bfseries Sum \\
\cmidrule(l@{\tabcolsep}r@{\tabcolsep}){2-3}
                & Train & Val &      \\
\midrule
  Causal        &   939 &  89 & 1028 \\
  Countercausal &   834 & 118 &  952 \\
  Uncausal      &  1302 & 133 & 1435 \\
\midrule
  Sum           &  3075 & 340 & 3415 \\
\bottomrule
\end{tabular}%
\hspace{2em}%
\begin{tabular}[t]{@{}lrrr@{}}
\addlinespace[-2ex]
(b)\\
\toprule
  \textbf{Property}   &     &     &       \textbf{Value} \\
\cmidrule(l@{\tabcolsep}){2-4}
                      & min & max &                  avg \\
\midrule
  Char length         &  11 & 908 &                174.7 \\
  Token length        &   2 & 136 &                 28.9 \\
\midrule
  Flesch Reading Ease &     &     &                   43 \\
\bottomrule
\end{tabular}%
\hspace{2em}%
\begin{tabular}[t]{@{}l@{}}
\addlinespace[-2ex]
(c)\\
\includegraphics[scale=0.75]{type-statistic.ai}
\end{tabular}\vspace{-1em}%
\caption{%
(a)~Corpus size and class distribution for the causality detection task.
(b)~Properties of the texts in the corpus.
(c)~Distribution of the different expressions of countercausal claims in the dataset.}
\label{table-corpus-statistics}
\end{table*}

%% file: concausality-part5.tex
\section{Experiments}
\label{sec:experiments}

This section presents baseline experiments on the \dataset{} (CCNC) and analyzes how current causality extraction models handle countercausal statements. The experiments
(1)~establish baselines for the newly introduced tasks, and
(2)~demonstrate that models trained without explicit supervision for countercausality misclassify countercausal claims as causal.

\paragraph{Setup}
We fine-tune three representative pre-train\-ed transformer models: DistilBERT \cite{sanh:2019}, RoBERTa \cite{liu:2019}, and Mis\-tral-7B-Instruct \cite{jiang:2023a}. DistilBERT and RoBERTa represent distilled and state-of-the-art bidirectional encoder-based architectures, respectively, while Mistral represents an autoregressive decoder-only large language model. All models are fine-tuned with a batch size of 256. Due to memory constraints, Mistral is trained using bfloat16 quantization.

The experiments follow the tasks introduced in Section~\ref{sec:tasks}: Task 1 is a binary sentence classification task, Task 2 is a sequence tagging task with BIO labels, and Task 3 is a ternary classification task. All tasks are evaluated using macro-averaged precision, recall, and $F_1$.

\paragraph{Baseline Results}
We first evaluate the models' effectiveness on the extended causality extraction tasks when trained on the \dataset{}. DistilBERT and RoBERTa are fine-tuned on all three tasks. Mistral is fine-tuned only on Tasks 1~and~3, as Task~2 requires token-level representations, which decoder-only transformers do not provide.

\input{concausality-table-corpus-evaluation.tex}

Overall, the results (Table~\ref{tbl:baseline}) show that pre-trained transformers are effective at causality extraction with an explicit countercausality label. RoBERTa consistently achieves the highest effectiveness across tasks, followed by DistilBERT, while Mistral performs substantially worse. These results provide the first baselines for the \dataset{} and serve as a point of comparison for future work on countercausality-aware extraction models.

\paragraph{Countercausality}
To investigate how models handle countercausal statements when not explicitly trained to recognize them, we additionally fine-tune each model on causality identification (Task~3) on the CNCv2, which does not have a specific label for countercausality. We then evaluate the trained models on all entity pairs of the \dataset{}.

\bsfigure[width=\linewidth]{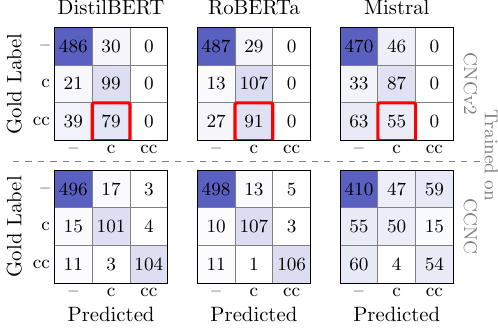}{Confusion matrices for the classification of entity pairs after training without (top) and with (bottom) explicitly handling countercausal claims. The class labels are no relationship (--), causal (c), and countercausal (cc). The red highlights mark countercausal claims misclassified as causal, which should be marked as ``--'' in absence of a countercausal label during training.}

Figure~\ref{confusion-matrices} shows the confusion matrices for each model. The top row presents results for models fine-tuned on the CNCv2. Since CNCv2 does not label countercausality explicitly, instances with this gold label (cc) should be predicted as ``no relationship'' (--). However, many of these instances are predicted as causal (c), which is highlighted in the confusion matrices.
All models exhibit a large number of such errors. Countercausal statements can closely resemble causal ones and therefore constitute hard negatives; for example, ``It is not that A causes B'' versus ``A causes B.'' 

As seen in the bottom row, models fine-tuned on the \dataset{} explicitly handle countercausal statements (ideal values outside the main diagonal are 0).
Without explicit supervision, the models in our experiments misclassified countercausal as causal over 10 times as often.

\paragraph{Discussion}
Table~\ref{tbl:baseline} and Figure~\ref{confusion-matrices} demonstrate that transformer-based models can effectively distinguish causal claims from countercausal claims when explicitly trained on such data. However, they tend to misclassify countercausality as causality when trained without explicit labels (as is the previous state of the art). In practice, such errors are severe, as misclassifying countercausality completely inverts the extracted meaning and invalidates any downstream causal reasoning. These findings add to the arguments presented in Section~\ref{sec:reasoning}: extraction of countercausal claims is not only necessary for reasoning but also important to eliminate a systematic failure mode.

Finally, although Mistral exhibits qualitatively similar trends to DistilBERT and RoBERTa, its overall performance is considerably worse. This result is somewhat surprising given its substantially larger parameter count and richer inherent world knowledge. One explanation is that bidirectional encoder-based models are better suited for causality detection, which relies on full-sentence context. To our knowledge, prior work has not applied autoregressive transformers to causality identification.

%% file: concausality-table-corpus-evaluation.tex
\begin{table}\centering\small%
\tabcolsep=1.931pt%
\newcommand{\rot}[1]{#1}
\newcommand\tmp[1]{\phantom{00}\makebox[0pt][r]{#1}}
\begin{tabular}{@{}l c@{~\,}c@{~\,}c @{~~} c@{~\,}c@{~\,} c@{~~} c@{~\,}c@{~\,}c @{}}\toprule
\textbf{Model} & \multicolumn{3}{c}{\textbf{Detection}} & \multicolumn{3}{c}{\textbf{Event Extr.}} & \multicolumn{3}{c@{}}{\textbf{Identification}}\\\cmidrule(lr){2-4}\cmidrule(lr){5-7}\cmidrule(l){8-10}
& $F_1$ & $P$ & $R$ & $F_1$ & $P$ & $R$ & $F_1$ & $P$ & $R$ \\
\midrule
\rot{DistilBERT}  &     \tmp{80}.0 &     \tmp{79}.9 &     \tmp{80}.0 &     \tmp{35}.8 &     \tmp{32}.4 &     \tmp{40}.1 &     \tmp{90}.1 &     \tmp{90}.7 &     \tmp{89}.5 \\
\rot{RoBERTa}     & \bf \tmp{87}.4 & \bf \tmp{87}.3 & \bf \tmp{87}.4 & \bf \tmp{44}.0 & \bf \tmp{41}.7 & \bf \tmp{46}.5 & \bf \tmp{92}.1 & \bf \tmp{92}.5 & \bf \tmp{91}.8 \\
\rot{Mistral}     &     \tmp{66}.2 &     \tmp{78}.7 &     \tmp{66}.4 &     \tmp{--}-- &     \tmp{--}-- &     \tmp{--}-- &     \tmp{56}.0 &     \tmp{56}.6 &     \tmp{55}.6 \\
\bottomrule
\end{tabular}\vspace{-1ex}
\caption{Macro-averaged $F_1$ scores, precision ($P$) and recall ($R$) of the baseline models on \dataset. All values are in percent (\%). The best values per metric are marked bold.}
\label{tbl:baseline}
\end{table}

%% file: concausality-sum.tex
\section{Conclusion}

Countercausal claims are necessary for reasoning on incomplete knowledge. However, we observe that models trained solely on causal claims tend to misclassify countercausal claims as causal. That is, statements claiming ``A~does not cause~B'' are extracted as ``A~causes~B'' or ignored.

To address this issue, we extend the causality extraction task to include countercausal claims and define and validate a detailed annotation guidelines to create the first dataset for training models on this extended task. The inclusion of this information during training is crucial to enable the correct handling of countercausal information in causality extraction. Our baseline results show that transformers are effective at distinguishing causality from countercausality.

\paragraph{Future Work}
Adding countercausality to causality research in natural language processing opens many new possibilities. Models that can extract causal and countercausal statements from text data can produce inconsistent information in causal graphs since it may contain the causal chain $A{\rightarrow}B{\rightarrow}C$ but also the countercausal relation $A{\not\rightarrow}C$, which needs to be resolved (e.g., by considering the supporting statements). This also poses an interesting new avenue for computational argumentation. Computational argumentation can be viewed as epi\-ste\-mic causality: \textit{Because \textlangle{}premises\!\textrangle{}, I believe \textlangle{}conclusion\!\textrangle{}}, where an argument can be attacked with countercausal claims as refuting evidence.

Finally, our dataset may be used to bootstrap further countercausal datasets, for example, by training models that generate countercausal texts from causal inputs, or training a classifier on our dataset and crawling for countercausality.

%% file: concausality-limitations.tex
\section*{Limitations}
The rephrased sentences may not be semantically correct. For example, the causality expressed in
\begin{center}\itshape
    \parbox{\dimexpr\linewidth-1cm\relax}{The workers went on strike five weeks ago demanding a minimum pay of R9000 a month.\linebreak[1]\null\hfill{\normalfont (\texttt{train\_04\_257\_234})}}
\end{center}
is rephrased to
\begin{center}\itshape
    \parbox{\dimexpr\linewidth-1cm\relax}{The workers went on strike five weeks ago \textbf{despite} demanding a minimum pay of R9000 a month.}
\end{center}
While this can be seen as a limitation, we believe that, in the contrary, it should not hinder a good model at picking up the countercausal nature of this statement since it remains clear. In the example above, it is clear that a countercausal relation is expressed between the demand for a certain minimum pay and the strike. This may even help discerning whether the models classify and extract (non)\nolinebreak{}causal relations correctly because they know the relation from their training data or because they can faithfully extract the information provided in the texts.

%% file: concausality-ethics.tex
\section*{Ethical Considerations}
The countercausal annotations negate causal sentences from news articles. As such they explicitly contain factually wrong information on actual events. We will make explicitly clear in the description accompanying the dataset that it must not be used to train factual information into a model and any information within the dataset must not be understood as truth.

\paragraph{Third Party Artifacts}
Our usage of the Causal News Corpus v2~\cite{tan2023a} adheres to its CC0 license\footnote{\href{https://github.com/tanfiona/CausalNewsCorpus/blob/master/LICENSE}{github.com/tanfiona/CausalNewsCorpus}} which allows us to modify and redistribute the dataset.

We cited the third party artifacts we used where appropriate in the paper. Beyond those, we also made use of the following frameworks: To annotate the dataset, we used Doccano~\cite{nakayama:2018}. We used the Transformer~\cite{wolf:2020}, Dataset~\cite{lhoest:2021}, and Evaluate frameworks by Hugging Face, PyTorch~\cite{ansel:2024}, Pandas~\cite{thepandasdevelopmentteam:}, and Jupyter Notebook~\cite{kluyver:2016} to train the baseline model and TIREx Tracker~\cite{hagen:2025b} for efficiency measurements and metadata.

%% file: concausality-appendix.tex
\section{Appendix}

\subsection{Terminology}
\label{sec:naming}

The following paragraphs list a few names we considered for countercausal claims before we settled with the term \emph{countercausal} to denote them. In all, \emph{noncausal} denotes statements that are not causal, \emph{countercausal} denotes statements that refute a causal relationship, and \emph{uncausal} are statements that are neither causal nor countercausal. As such, countercausal and uncausal form a partition of noncausal statements.

\paragraph{Noncausal}
We believe that, morphologically, \emph{noncausal statement} would be the most fitting name for countercausal statements. Take for example the terms \emph{deterministic}, \emph{nondeterministic}, and \emph{not deterministic}. Here, nondeterministic is a separate class to deterministic but not the complement. For example, \emph{randomized} algorithms are not deterministic but not nondeterministic. Similarly, causal counterclaims could be called \emph{noncausal} and statements like ``\emph{He ate}'' would be considered neither \emph{causal} nor \emph{noncausal}. However, prior works unanimously used \emph{noncausal} synonymously with \emph{not causal}.

\paragraph{Acausal and Anti-causal}
In Physics, \emph{acausal} / \emph{anticausal} describe a dependency on time. E.g., the sentence ``If the balls' final state were different, their initial state would have to have been different'' is anti-causal / acausal~\cite{karimi:2003,frisch:2023}.

\paragraph{Negative causation}
The term \emph{negative causation} is already known in philosophy to denote that an event causes the inhibition of another event~\cite{bisketzis:2008}. As such, it is a subclass of the causal case. Note that, by contradiction, if A prevents B, then A cannot cause B. But there may be events A and B such that A neither causes nor prevents B. As such, negative causation is stronger than causal counterclaims.

\paragraph{Concausal}
From the prefix ``con'' meaning ``together''; two events jointly cause the effect.

\newpage
\subsection{Prompt-based Reformulation Test}
\label{sec:prompt-reformulation-test}

\begin{figure}[ht]
\begin{center}
\parbox{.95\linewidth}{\ttfamily\hyphenchar\font=\defaulthyphenchar\relax
You will be given a sentence that contains a causal statement. Your task is to identify the causal statement and negate it. Change as few words as necessary. Repeat the entire sentence but replace the causal statements with its negation.

The ANC in KwaZulu-Natal strongly condemns the misbehaviour of IFP members who interrupted our campaign trail led by ANC Deputy President Jacob Zuma at Dokodweni and Mandeni on the north coast today
}
\end{center}
\caption{Example prompt of our preliminary experiments on rewriting causal statements to be countercausal using GPT-4.}
\label{fig:example_prompt}
\end{figure}

\input{concausality-table-gpt-example}

\clearpage
\onecolumn
\subsection{Annotation Guidelines}
\label{sec:annoation-guidelines}

\begin{figure*}[ht]
\centering
\includegraphics*[width=\linewidth]{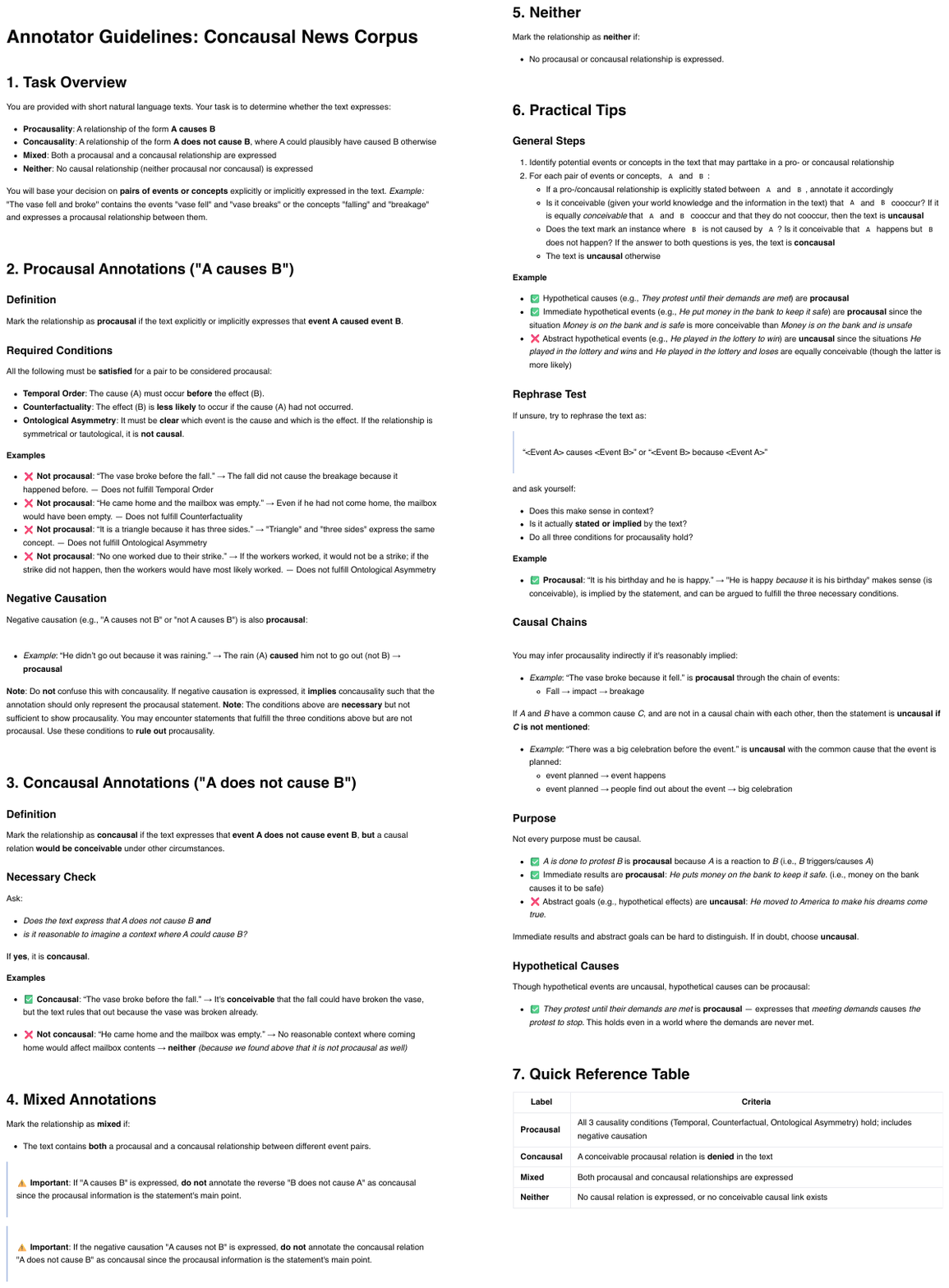}
\caption{The instructions given to the annotators for annotating our dataset.}
\label{fig:annot-guides}
\vspace{-15ex}
\end{figure*}

\clearpage
\onecolumn
\subsection{Excerpt of Training Data}
\label{sec:training-data-excerpt}

\input{concausality-table-misannotated}

%% file: concausality-table-gpt-example.tex
\begin{table}[ht]
\centering\small
\renewcommand{\arraystretch}{.85}%
\belowrulesep=\dimexpr\arraystretch \belowrulesep\relax%
\aboverulesep=\dimexpr\arraystretch \aboverulesep\relax%
\belowbottomsep=\belowrulesep%
\newcommand\rot[1]{\rotatebox[origin=c]{90}{#1}}
\renewcommand\hl[1]{\textcolor{black}{\bfseries #1}}
\begin{tabular}{@{}l@{\hspace{5pt}}>{\color{gray}}m{4.005cm}@{\hspace{7pt}}c@{}}\toprule
    \multicolumn{2}{@{}l}{\textbf{Sentence}} & \makecell[t]{\textbf{Expressed}\\\textbf{(Counter-)claim}}\\\midrule
    \rot{Orig.} & \hl{The ANC in KwaZulu-Natal strongly condemns the misbehaviour of IFP members who interrupted our campaign trail} & $\textrm{interrupt} \to\textrm{condemn}$\\\noalign{\vskip 5pt}
    \rot{Manual} & The ANC in KwaZulu-Natal \hl{did not condemn} the misbehaviour of IFP members \hl{despite their interruption of} our campaign trail & $\textrm{interrupt} \not\to\textrm{condemn}$\\\noalign{\vskip 5pt}
    \rot{GPT-4o} & The ANC in KwaZulu-Natal strongly condemns the misbehaviour of IFP members who \hl{did not interrupt} our campaign trail & $\lnot\textrm{interrupt} \to\textrm{condemn}$\\\bottomrule
\end{tabular}
\caption{Different reformulations of training sample \texttt{train\_06\_60\_1969} from CNCv2. Instead of producing a countercausal claim, GPT-4o incorrectly negated the cause, resulting in a causal claim. GPT-4o's response was obtained using the prompt in Figure~\ref{fig:example_prompt}.}
\label{tbl:gptexamples}
\end{table}

%% file: concausality-table-misannotated.tex
\begin{table*}[ht]
\centering
\small
\begin{tabular}{@{}lp{10.991cm}l@{}}\toprule
\textbf{Identifier}            & \textbf{Text} & \textbf{Label}\\\midrule
\ttfamily train\_06\_304\_2720 & Speakers at the rally , orgainsed by the Peoples Union for Civil Liberties ( PUCL ) , CPI , CPM , and Chhattisgarh Mukti Morcha ( CMM ) , accused the Raman Singh government of having implicated Sen in a false case . & Uncausal \\
\ttfamily train\_07\_209\_2356 & Monday saw the continuing trend of protests in the city , as more than 500 people gathered at Town Hall .                                                                                                               & Uncausal \\
\ttfamily train\_08\_243\_130  & Will the unprecedented protests embolden them to fight for their beliefs in future , or convince them that resistance to Beijing ' s will is futile ?                                                                   & Uncausal \\
\ttfamily train\_08\_45\_984   & Similarly , some palmyrah farmers tapped toddy at Pattankaadu even as the police arrested 517 protestors , including 66 women , in the neighbouring town of Vasudevanallur .                                            & Countercausal \\
\ttfamily train\_08\_5\_2154   & PTI Guwahati Police Commissioner Mukesh Aggarwal said that the anti-talk faction of ULFA may be behind the attack .                                                                                                     & Uncausal \\
\ttfamily train\_05\_187\_787  & The police charged Mr. Chandrashekhar with instigating violence on May 9 under various IPC sections .                                                                                                                   & Uncausal \\
\ttfamily train\_06\_249\_2072 & The protesters raised slogans against the government .                                                                                                                                                                  & Uncausal \\
\ttfamily train\_08\_67\_544   & " We had gone to study the life of people in remote and Naxal-affected tribal areas as part of our mission and did not expect to be kidnapped by the Naxals , though we fully knew about their presence , " they said . & Countercausal \\
\ttfamily train\_08\_76\_2969  & Organisers said almost 300,000 protesters and residents on Saturday afternoon defied a police ban to descend on the town in Hong Kong ' s western New Territories .                                                     & Countercausal \\
\ttfamily train\_05\_39\_3240  & Despite the march being peaceful , most of the businesses in the inner city were closed .                                                                                                                               & Countercausal \\
\ttfamily train\_07\_11\_827   & Vehicles have also been used to commit attacks on civilians in Nice and Berlin .                                                                                                                                        & Uncausal \\\bottomrule
\end{tabular}
\caption{Excerpt of training samples from the Causal News Corpus v2 that were wrongly annotated as causal.}
\label{tbl:misannotated}
\end{table*}